\documentclass[10pt,a4paper]{article}
\usepackage{lrec}
\usepackage{graphicx}
\usepackage{tabularx}
\usepackage{soul}
\usepackage[latin1]{inputenc}
\usepackage{enumitem}   
\usepackage[hidelinks]{hyperref}
\usepackage{xstring}
\usepackage{listings}
\usepackage{xcolor}

\usepackage{titlesec}
\titleformat{\section}{\normalfont\large\bf\center}{\thesection.}{1em}{}
\titleformat{\subsection}{\normalfont\SmallTitleFont\bf\raggedright}{\thesubsection.}{1em}{}
\titleformat{\subsubsection}{\normalfont\normalsize\bf\raggedright}{\thesubsubsection.}{1em}{}
\renewcommand\thesection{\arabic{section}}
\renewcommand\thesubsection{\thesection.\arabic{subsection}}
\renewcommand\thesubsubsection{\thesubsection.\arabic{subsubsection}}

\colorlet{punct}{red!60!black}
\definecolor{background}{HTML}{EEEEEE}
\definecolor{delim}{RGB}{20,105,176}
\colorlet{numb}{magenta!60!black}

\lstdefinelanguage{json}{
    basicstyle=\small\ttfamily,
    numbers=left,
    numberstyle=\scriptsize,
    stepnumber=1,
    numbersep=8pt,
    showstringspaces=false,
    breaklines=true,
    frame=lines,
    backgroundcolor=\color{background},
    literate=
     *{0}{{{\color{numb}0}}}{1}
      {1}{{{\color{numb}1}}}{1}
      {2}{{{\color{numb}2}}}{1}
      {3}{{{\color{numb}3}}}{1}
      {4}{{{\color{numb}4}}}{1}
      {5}{{{\color{numb}5}}}{1}
      {6}{{{\color{numb}6}}}{1}
      {7}{{{\color{numb}7}}}{1}
      {8}{{{\color{numb}8}}}{1}
      {9}{{{\color{numb}9}}}{1}
      {:}{{{\color{punct}{:}}}}{1}
      {,}{{{\color{punct}{,}}}}{1}
      {\{}{{{\color{delim}{\{}}}}{1}
      {\}}{{{\color{delim}{\}}}}}{1}
      {[}{{{\color{delim}{[}}}}{1}
      {]}{{{\color{delim}{]}}}}{1},
}

\title{Orchestrating NLP Services for the Legal Domain}

\name{\hspace*{-3mm}\parbox{\textwidth}{\centering 
Juli\'{a}n Moreno-Schneider\textsuperscript{1}, 
Georg Rehm\textsuperscript{1}, 
Elena Montiel-Ponsoda\textsuperscript{2},\\
V\'{i}ctor Rodr\'{i}guez-Doncel\textsuperscript{2},
Artem Revenko\textsuperscript{3},
Sotirios Karampatakis\textsuperscript{3},\\
Maria Khvalchik\textsuperscript{3},
Christian Sageder\textsuperscript{4},
Jorge Gracia\textsuperscript{5}, 
Filippo Maganza\textsuperscript{6}
}\vspace*{2mm}}

\address{
\textsuperscript{1} DFKI GmbH, Germany --
\textsuperscript{2} Universidad Polit\'{e}cnica de Madrid, Spain -- 
\textsuperscript{3} Semantic Web Company GmbH, Austria \\
\textsuperscript{4} Openlaws GmbH, Austria --
\textsuperscript{5} Universidad de Zaragoza, Spain --
\textsuperscript{6} Alpenite, Italy \\[1ex]
    Corresponding author: Juli\'{a}n Moreno Schneider -- julian.moreno\_schneider@dfki.de\\}

\abstract{Legal technology is currently receiving a lot of attention from various angles. In this contribution we describe the main technical components of a system that is currently under development in the European innovation project Lynx, which includes partners from industry and research. The key contribution of this paper is a workflow manager that enables the flexible orchestration of workflows based on a portfolio of Natural Language Processing and Content Curation services as well as a Multilingual Legal Knowledge Graph that contains semantic information and meaningful references to legal documents. We also describe different use cases with which we experiment and develop prototypical solutions.
\\ \newline  
}

\begin{document}

\maketitleabstract

\section{Introduction}
\label{sec:introduction}

We present a methodology and tooling to handle a set of various Natural Legal Language Processing and Document Curation services currently under development in the EU project Lynx\footnote{\url{http://lynx-project.eu}}. First, the platform is acquiring data and documents related to compliance from jurisdictions in different languages with a focus on English, Spanish, German, and Dutch along with terminologies, dictionaries and other language resources. Based on this collection of structured data and unstructured documents we create a multilingual Legal Knowledge Graph (LKG), represented as Linked Data. Second, a set of flexible language processing services is developed to analyse and process the data and documents to integrate them into the LKG. Semantic processing components annotate, structure, and interlink the LKG contents.

The following research challenges arose during the project:
\begin{enumerate}[label=(\roman*)]
\item How to efficiently orchestrate a set of NLP services? How to guarantee their interchangeability? \label{rq1}
\item How to efficiently extract information and store documents along with the extracted information? \label{rq2}
\item In which business scenarios are legal NLP services able to generate actual added value? \label{rq3}
\end{enumerate}

The remainder of this article is structured as follows. Section~\ref{sec:usecases} describes different use cases, addressing challenge~\ref{rq3}, while Section~\ref{sec:lkg} focuses upon the LKG used in the prototype applications, addressing challenge~\ref{rq2}. The infrastructure, semantic services and their orchestration through the content and document curation workflow manager, is described in Section~\ref{sec:platform}, addressing challenge~\ref{rq1}. After a brief review of related work (Section~\ref{sec:relatedwork}) we summarise the paper and describe future work (Section~\ref{sec:conclusions}).

\section{Use Cases} \label{sec:usecases}

The following three briefly sketched use cases illustrate the development work in the project. The pilot initiatives described next are still not operational, but the current efforts will make them operational in the summer of 2020.

The objective of the \emph{Contract Analysis} use case is to enhance regulatory compliance and obligations through automation, reducing costs, corporate risks and personal risks. Currently, companies have to manage large amounts of heterogeneous contracts, which is, typically, a time-consuming manual process. SMEs do not use management systems to help them in identifying the core data and main actions enforced by a contract. Usually, only a minimal amount of information is kept in a spreadsheet (title, parties, date of signature), which is insufficient to effectively manage contracts and keep track of the actions that need to be taken by the company (such as renewal or amendments). To ensure compliance, improve governance and mitigate risks, companies need to rely on systems that support them in digitizing contracts, identifying the core data, providing an overview of the main content, pointing to the relevant legislation in force, and sending out notifications in case actions need to be taken. Some work has already been done in this sense \cite{Chalkidis2017ADL}. The prototype will extract information from contracts and subsequently monitor and analyze the documents against (a) the public regulatory framework (including legislation and case law from the EU and Member States, public provisions and suggestions by authorities, etc.) and (b) private contracts. The system will actively inform companies, persons in charge (directors, managers, data protection officers, etc.) or the company's lawyer(s) whenever there are updates in relevant legislation, case law, or in contractual obligations that affect a company's obligations, even across different jurisdictions and languages.

The \emph{Labour Law} use case provides access to aggregated and interlinked legal information regarding labour law across multiple legal orders, jurisdictions, and languages. The prototype analyses labour legislation from the EU and Member States, and jurisprudence related to labour law issues. This use case makes use of lists of Frequently Asked Questions (FAQ) regarding employment and labour relations, which should be privileged whenever looking for answers posed in natural language. 
The platform addresses the integration of these heterogeneous documents, coming from different jurisdictions, in various languages, with unequal structure, temporal validity, and geographical scope, which will ultimately benefit the Digital Single Market. 

The \emph{Oil and Gas} use case is focused on compliance management support for geothermal energy projects and aims to obtain standards and regulations associated with certain terms in the field of geothermal energy. A user can submit a Request For Proposal (RFP) or feasibility study to the system and is then informed about which standards or regulations must be taken into consideration to carry out the considered project in a compliant manner. This scenario will innovate and speed up existing compliance related services. The uploaded user documents (RFP or feasibility study) are analyzed and, with the help of semantic services (Section~\ref{sec:platform}), the most important terms are identified. The terms of interest include geolocations, types of activities, types of machinery involved, names of organizations, possible mentions of relevant regulations. Next up, the prototype uses the search service to find the documents in the LKG that are most relevant to the uploaded user document. The ranking of the retrieved documents is executed by the semantic similarity service. Thanks to machine translation, the prototype is able to deal with multiple languages -- the documents presented to the user are not necessarily in the same language as the uploaded document.

\section{Legal Knowledge Graph}
\label{sec:lkg}

Knowledge graphs represent entities as nodes, attributes as node labels and the relationship between entities as edges. RDF is particularly well suited for representing knowledge graphs, and, indeed, the recent attention has finally brought Semantic Web technologies back into the centre of current research and development trends after years of silent existence. Many AI and NLP applications rely on knowledge graphs as crucial resources, such as, among others, information search, data integration, data analytics, question answering and context-dependent recommendations.  

In the multilingual legal domain, knowledge graphs have the full support of public institutions, which are publishing massive amounts of linked data, which are becoming critical assets of the companies operating them. The amount of legal data made accessible either in free or for-a-fee modalities by legal information providers can be hardly imagined. In 2014, Lexis Nexis claimed to have 30 Terabytes of content, WestLaw accounted for more than 40,000 databases\footnote{LexisNexis Legal and Professional, see \url{http://lexisnexis.com}}. Their value can be roughly estimated: as early as 2012, the four big players (WestLaw, Lexis Nexis, Wolters Kluwer and Bloomberg Legal) about US-\$10,000M in total revenue. Language data (e.\,g., resources with any kind of linguistic information) belongs to a much smaller domain, but is still, unmanageable as a whole.

\begin{figure*}[h]
\centering
\includegraphics[width=.65\textwidth]{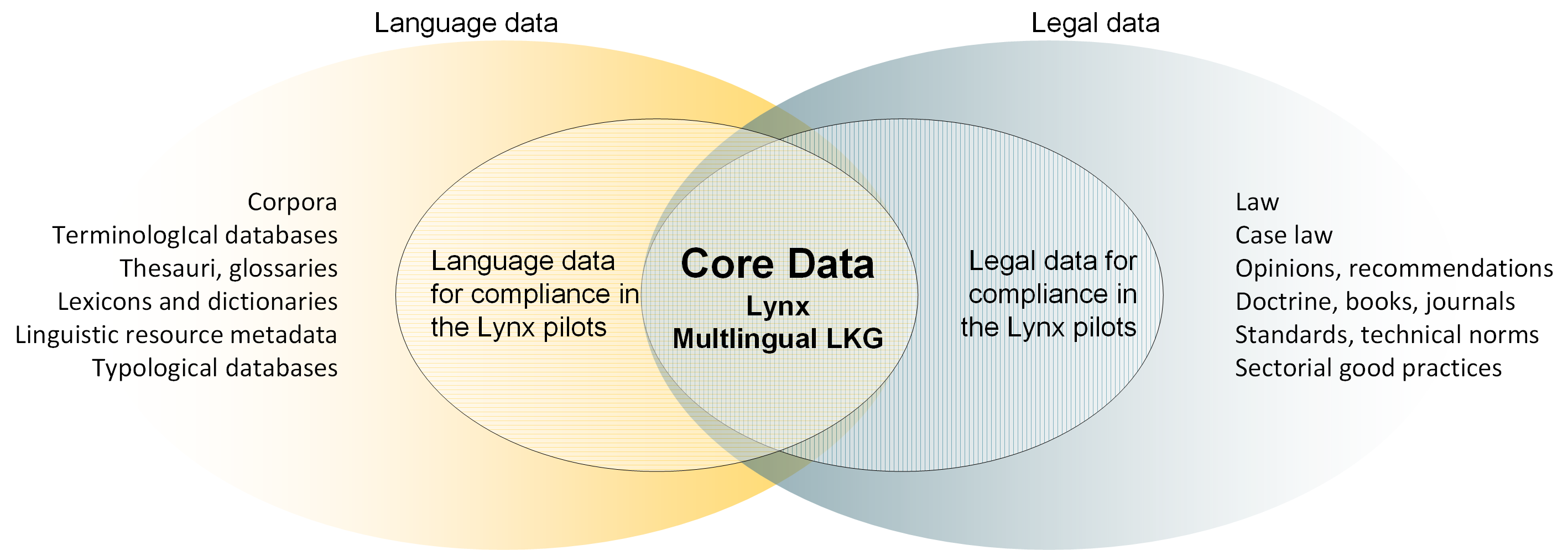}
\caption{Scope of the Core Data in the Multilingual Legal Knowledge Graph}
\label{figlkg1}
\end{figure*}

We are interested in a small fraction of the information belonging to these domains. In particular, Lynx is using the data necessary to provide the compliance services (Section~\ref{sec:usecases}) -- the earliest Lynx knowledge graph being the Spanish legislation enriched with annotations \cite{spanishlegislation}. As shown in Figure~\ref{figlkg1}, the scope of the data in the Lynx Multilingual Legal Knowledge Graph is legal and regulatory data (mainly comprising legislation, case law and standards-related data), on the one hand, and language data (such as corpora, terminologies, glossaries or dictionary data), on the other, to cover the multilingual aspects of the services. The Lynx platform will try to comprehensively identify every possible open dataset in the intersection of these domains as its core category. 

Figure~\ref{figlkg1} shows the core of the Lynx Multilingual Legal Knowledge Graph in the strict sense, as the set of entities whose URI is within the `lynx' top level domain, but with links to external entities in the wider Web of Data. The definitions of both language data and regulatory data are fuzzy, but flexible as to introduce data of many different kinds whenever necessary (geographical data, user information, etc.). Because data in the Semantic Web can not be separated from the data models, and data models are accessed in the same manner as data, ontologies and vocabularies are part of the LKG as well. Moreover, any kind of metadata (describing documents, standards etc.) is also part of the LKG, as well as the description of the entities producing the documents (courts, users, jurisdictions). In order to provide the compliance services, and with different degree of interest, both primary and secondary law are of use, and any relevant document in a wide sense may become part of the Legal Knowledge Graph.

\section{Lynx Platform}
\label{sec:platform}

The base infrastructure of the Lynx platform follows the paradigm of a microservice architecture, which is a variant of the service-oriented architecture (SOA) where an application is structured as a collection of loosely coupled services. Each microservice can be developed and deployed independently, which also allows the use of different programming languages for their implementation. Microservices are small and autonomous and can be developed more efficiently than monolithic, integrated systems. In addition, the deployment of microservices can, to a very large extent, be automated, also facilitating the monitoring of individual services. A crucial advantage is concerned with the scalability of systems based on microservices, which is a lot easier than scaling monolithic systems. The communication between services is based on REST interfaces, allowing the simple communication and decoupling of the client from the server. We decided to make use of containers (specifically Docker\footnote{\url{https://www.docker.com}}) and an architecture that can host and manage several containers (OpenShift\footnote{\url{https://www.openshift.com}}, a containerization software built on top of Kubernetes\footnote{\url{https://kubernetes.io}}).

\subsection{Semantic Services}
\label{ssec:services}

In the following we describe the semantic processing services. 
This is a heterogeneous set: some of the services make use of others, some extract or annotate information (e.\,g., NER or Temporal Expression Analysis), while others operate on full documents, yet others provide a user interface (e.\,g., QA). A complete description is out of scope for this paper and can be found in \newcite{rehm2019c}.

\paragraph{Term Extraction} Enables the creation of a taxonomy for a certain use case, domain or company, using the cloud-based Tilde~Terminology service\footnote{\url{https://term.tilde.com}}. It extracts terms following the methodology by \cite{Pinnis2012}, creating a SKOS vocabulary containing terms, contexts and references to their source documents.

\paragraph{Linguistic Resources} The LKG has to be adaptable across domains and sectors. It is based on a collection of domain-dependent and domain-independent vocabularies accessible through a common RDF graph. The domain-dependent vocabularies comprise terminologies coming from the legal sector and the use case domains (e.\,g., EuroTermBank\footnote{\url{http://www.eurotermbank.com}}), and others created from scratch to cover the specific needs of the business cases, taking advantage of the Linguistic Linked Open Data Cloud (LLOD). The domain-independent vocabularies are taken from lexicographic data published by KDictionaries.\footnote{\url{https://www.lexicala.com}} They contain cross-lingual links for the five languages served by our platform (Dutch, English, German, Italian, Spanish). Besides their overall coverage of solely domain-independent vocabularies, they contain information on words and phrases that include also or only domain-dependent meanings (e.\,g., court for the former, lawyer for the latter). Domain-independent dictionary data provide a common ground across domains that facilitates traversing semantically annotated documents coming from different specialised domains (e.\,g., Legal or Oil \& Gas). They also support certain NLP functionalities such as Word Sense Disambiguation by providing a common catalogue of word senses. 
The data is being remodeled in RDF according to the Ontolex Lemon Lexicography Module Specification.\footnote{\url{http://www.w3.org/ns/lemon/lexicog}} 
Right now that linguistic information is being used by other services, WSD, Search, QADoc, to get synonyms, term variants and translations that help in the cross-lingual search and cross-lingual question answering.

\paragraph{Named Entity Recognition} The NER service \cite{rehm2019d,leitner2020} is based on several trained models, Conditional Random Fields (CRFs) and bidirectional Long-Short Term Memory Networks (BiLSTMs) \cite{huang2015bidirectional,lample2016neural,riedl2018named}. 
This service includes an entity linking module in which we retrieve a unique identifier (URI) for the spotted entities. It uses DBPedia SPARQL\footnote{\url{https://dbpedia.org/sparql}} and DBPedia spotlight.\footnote{\url{https://www.dbpedia-spotlight.org}} Retrieved URIs are stored as part of the entity annotation. 

\paragraph{Concept Extraction} Concept extraction enables the insertion of links between documents and elements of controlled vocabularies in the LKG. These relations are the first step for enriching text fragments with knowledge from the LKG. Importantly, the inclusion of labels in many languages allows linking of documents in different languages, combining the knowledge derived from them, as well as multilingual search and recommendation. 
The Concept Extraction service works in as many languages as the taxonomies have labels in, and thus we can leverage multinational efforts for creating multilingual taxonomies such as EUROVOC\footnote{\url{https://publications.europa.eu/en/web/eu-vocabularies/}} or UNBIS\footnote{\url{http://metadata.un.org/?lang=en}}. 

\paragraph{Word Sense Disambiguation} To enable the use of incomplete KGs for automatic text annotations, we introduce a robust method for discriminating word senses using thesaurus information like hypernyms, synonyms, types/classes, contained in the KG \cite{Revenko2017DiscriminationOW}. It uses collocations to induce word senses and to discriminate the thesaurus sense from others. The service is used for any kind of entity linking, especially after NER, to correctly identify which named entities are indeed within the vocabulary scope of the LKG.

\paragraph{Temporal Expression Analysis} A prototype analyses time expressions in German-language legal documents, especially court decisions and legislative texts. The annotation of temporal expressions is important, but the most interesting part is the normalization, which can be used for interlinking documents (or parts of documents) using the normalized values of temporal expressions.

\paragraph{Legal Reference Resolution} Usually, editors attempt to be consistent and follow patterns to reference other documents. The developed methodology, currently implemented as a language-agnostic prototype, follows this assumption and attempts to discover patterns used in a semi-automatic manner. Patterns are constructed from features that are either individual tokens (e.\,g., ``Decision'', ``EU'', etc.) or processed features (e.\,g., ``DIGITS'' as a placeholder for numbers). 

\paragraph{Semantic Similarity} We use a hybrid type of similarity measure. First, the text of the document is annotated, such as the resolution of temporal or geographical references. Second, similarity is computed using a linear combination of text-based and knowledge-based similarities. The former are encoded by cosine-similarity of TF-IDF vectors and the latter by the overlap as measured by Jaccard coefficient of entities that the two documents either mention directly, or are linked in the LKG to mentioned ones. This approach allows us to detect similarity between documents even if they have only few entities in common. 

\paragraph{Question Answering} The Question Answering (QA) service accepts a natural language question and responds with an answer, extracted from a document in a given corpus. The end-to-end system consists of three components: 1) The Query Formulation module transforms a question into a query, which can be expanded using a domain specific vocabulary from the LKG. The query is processed through an indexer to obtain matching documents from the corpora. 2) The Answer Generation module extracts potential answers from the retrieved documents from the LKG. 3) The Answer Selection module identifies the best answer based on various criteria such as local structure of the text and global interaction between each pair of words based on specific layers of the model. 

\subsection{Document Manager}
\label{ssec:documentmanager}

The Document Manager (DCM) forms a central part of the Lynx platform in terms of the general platform capabilities; this is where documents are stored and maintained. Its basic functionality includes the storage and annotation of documents with an emphasis on keeping them synchronized, providing read and write access, as well as updates of documents and annotations. 

The DCM can be queried in terms of annotations (e.\,g., ``which documents contain mentions of this entity?''), and in terms of documents (e.\,g., ``what are the contents/annotations of document X?''). All queries to the DCM are executed through REST. The interface includes a set of Create, Read, Update, and Delete (CRUD) APIs to manage collections, documents and annotations within the Lynx platform.

Through their representation in JSON-LD (namely, RDF), Lynx documents are not only isolated elements but nodes in a graph. The use of semantics to formalize the meaning of the classes and properties qualifies this graph to be called an actual Knowledge Graph. The DCM is implemented as a Linked Data Platform (LDP) server based on Trellis\footnote{\url{https://github.com/trellis-ldp/trellis}}, which is why basic metadata about a document is stored as triples natively -- our implementation is based on Elastic Search and stores a JSON-LD serialisation of RDF. Document structure information and various types of metadata such as subject, jurisdiction, language etc.~are also triplified through the DCM at storing time. The NIF \cite{hellmann2012nif} Version 2.1 ontology is used to describe the structure metadata and a mashup of metadata-specific ontologies are used for other descriptive, structural or administrative metadata. Annotations of each document are also described using NIF V2.1. An overview of the Lynx data model can be found online.\footnote{\url{http://lynx-project.eu/data2/data-models}} Triples from all documents including data and metadata can be queried using the SPARQL endpoint provided by Trellis, thus providing access to the LKG including the ability to evaluate complex queries -- the equivalent for the ElasticSearch implementation being made by periodic data exports, queryable through the endpoint.\footnote{\url{http://sparql.lynx-project.eu}} Extensive usage of vocabularies as values for metadata or annotations increases the value of the LKG and the interoperability of the system. The DCM is the main building block of the Lynx Legal Knowledge Graph (LKG), it is where the LKG resides. Its basic architecture and core functionalities are described in \cite{lynxd14,lynxd41}. 

\subsection{Workflow Manager}
\label{ssec:workflowmanager}

\begin{figure*}[htbp]
\centering
\includegraphics[width=0.7\textwidth]{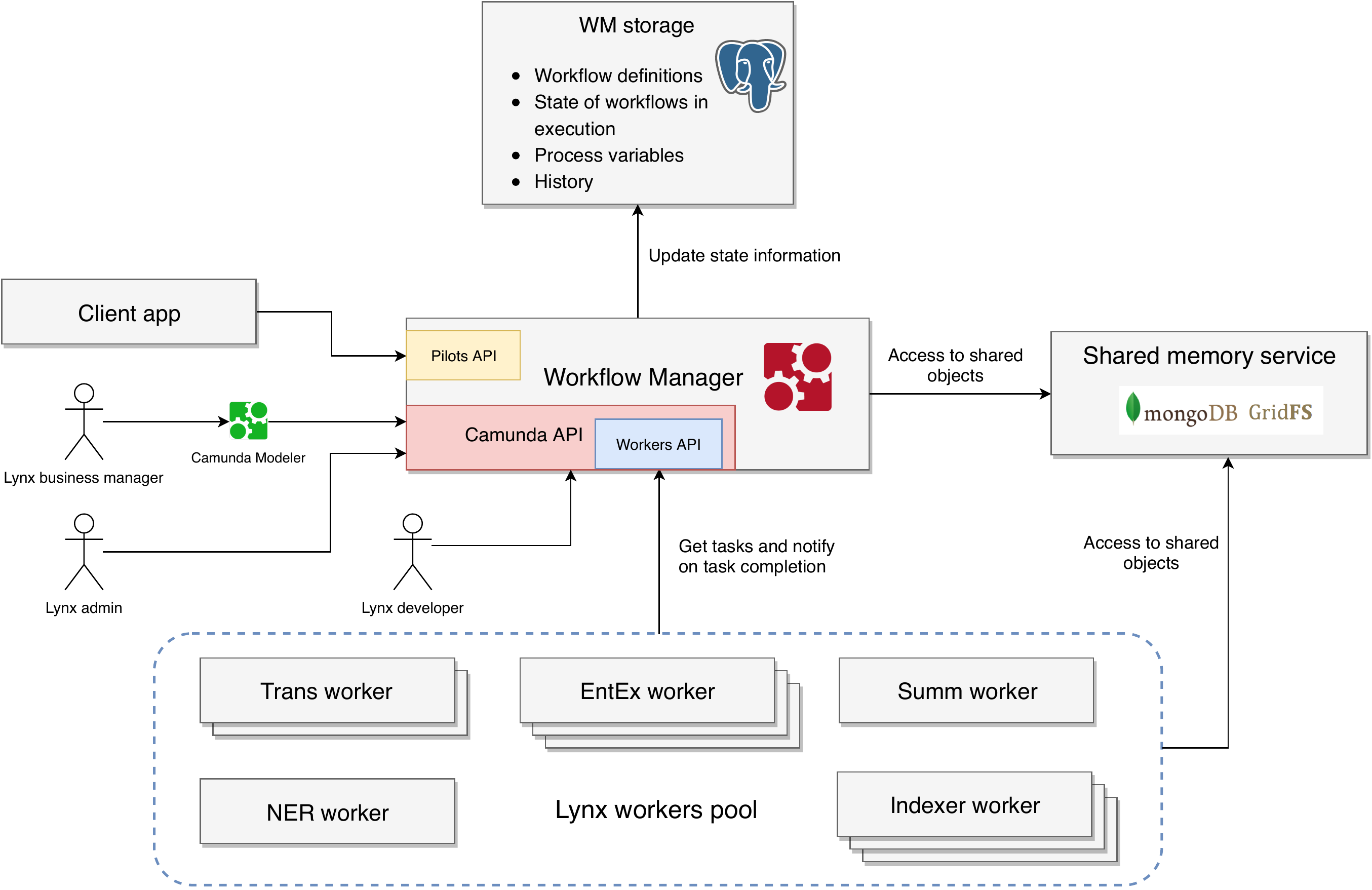}
\caption{Workflow Manager Architecture}
\label{fig:wmarchitecture}
\end{figure*}

Using a microservice architecture enforces the use of some kind of management tool in order to orchestrate the execution of the different services involved in more complex tasks \cite{lynxd14}. The combination of several functionalities from different services is defined as a workflow and the module responsible for orchestrating them is called workflow manager (WM). Our previous work includes a generic workflow manager for curation technologies \cite{rehm2016j}, and two indicative descriptions of the initial prototype of the Lynx workflow manager \cite{rehm2018g,rehm2018f}. The final Lynx workflow manager is based on the Camunda BPMN engine\footnote{\url{https://camunda.com}} because Camunda was in a more mature state than any of the alternatives. The requirements of the WM are presented in \cite{lynxd41}, while the Lynx workflows are described in \cite{lynxd42,lynxd43}. Figure~\ref{fig:wmarchitecture} shows the architecture of the workflow manager. Its main components are described in the following sections.

\subsubsection{Workflow Manager Engine}

The Workflow Manager Engine (WME) is responsible for converting workflows into tasks for the workers. It is based on Camunda. The main concepts of this component are:

\begin{itemize}
\item Workflow: a direct acyclic graph whose nodes are associated with tasks;
\item Task: an atomic unit of business logic, a task is associated with one and only one Lynx peripheral service;
\item Process: a runtime instance of a workflow;
\item Job: a runtime instance of a task.
\end{itemize}

The WME also provides a complete REST interface for managing workflow executions and templates. Apart from that, the WME uses some internal storage to store the different objects (workflows, tasks, processes and jobs) that are created during execution. All these elements are stored in the Workflow Manager Storage, which is implemented using a PostgreSQL\footnote{\url{https://www.postgresql.org}} database.

\subsubsection{Workers}

The Workers are responsible for the execution of tasks inside a workflow. Every task is identified by a topic name like ``TimEx-LKGPopulation'' or ``NER-ContractAnalysis''. Every worker uses the Camunda External Task Client library\footnote{\url{https://docs.camunda.org/manual/7.9/user-guide/ext-client/}} to connect to the WME to obtain the tasks it has to execute. Currently, there are four implemented types of workers. Each type can be instantiated multiple times with different configuration files, i.\,e., each instantiated worker is responsible (based on the configuration) to connect to a different service, or even to the same service but with different parameters (e.\,g., ``lang=de'' or ``lang=en''). The four types of workers have different functionalities: 

\begin{enumerate}[label=(\roman*)]
\item the \emph{document-translation-worker} connects to the Tilde translation services;
\item the \emph{document-enrichment-worker} connects to one of the enrichment services inside the Lynx platform (NER, TIMEX, SUMM, WSID, EntEx, etc.); 
\item the \emph{save-enriched-doc-in-LKG-worker} saves an enriched document inside the DCM; and
\item the \emph{create-enriched-document-worker} creates the enriched document. It collects all annotations produced by the services and aggregates them to create a Lynx document. This document is stored in the DCM or sent back to the client.
\end{enumerate}
 
\subsubsection{Shared Memory Service}

The shared memory service is used by both the workflow manager engine and the workers to share large data objects. For instance, the WM uses it to share with the workers the documents they have to process. The shared memory service uses a MongoDB\footnote{\url{https://www.mongodb.com}} database.

\subsubsection{Pilot API}

The Pilot API is a component of the WM and responsible for accessing, managing and executing the workflows of the project pilots. It consists of four discrete methods: three HTTP POST methods to execute and manage specific workflows (one method per pilot); and one HTTP GET method to retrieve the current state of a concrete workflow.

\subsubsection{Graphical User Interface}

Workflows are described using the BPMN (Business Process Model and Notation) standard \cite{omg2011bpmn}. Considering that specifying BPMN files manually is not the most user friendly approach, we integrated a graphical user interface for the definition of new workflows. We decided to use the Camunda Modeler.\footnote{\url{https://camunda.com/download/modeler/}} 

\subsection{Defined Workflows}
\label{ssec:implementedworkflows}

\begin{figure*}[tb]
\centering
\includegraphics[width=0.9\textwidth]{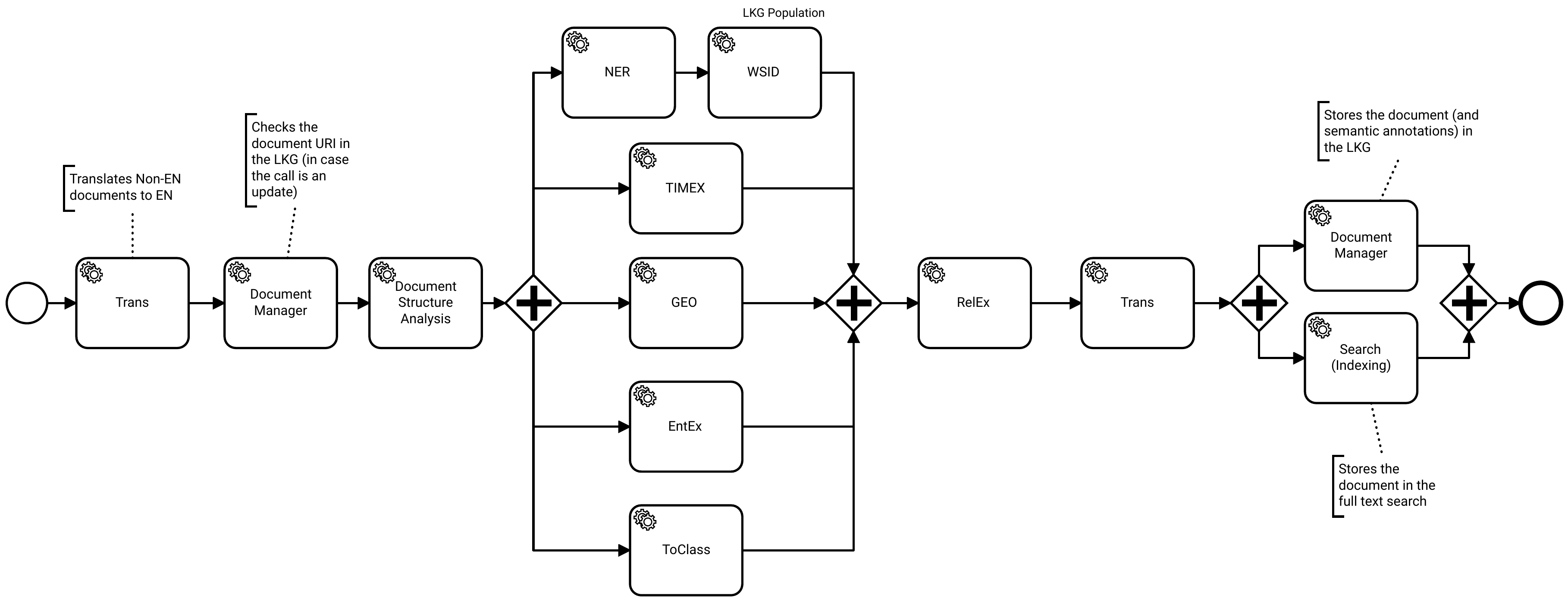}
\caption{Legal Knowledge Graph population workflow}
\label{fig:workflow-1}
\end{figure*}

\begin{figure*}[tb]
\centering
\includegraphics[width=0.78\textwidth]{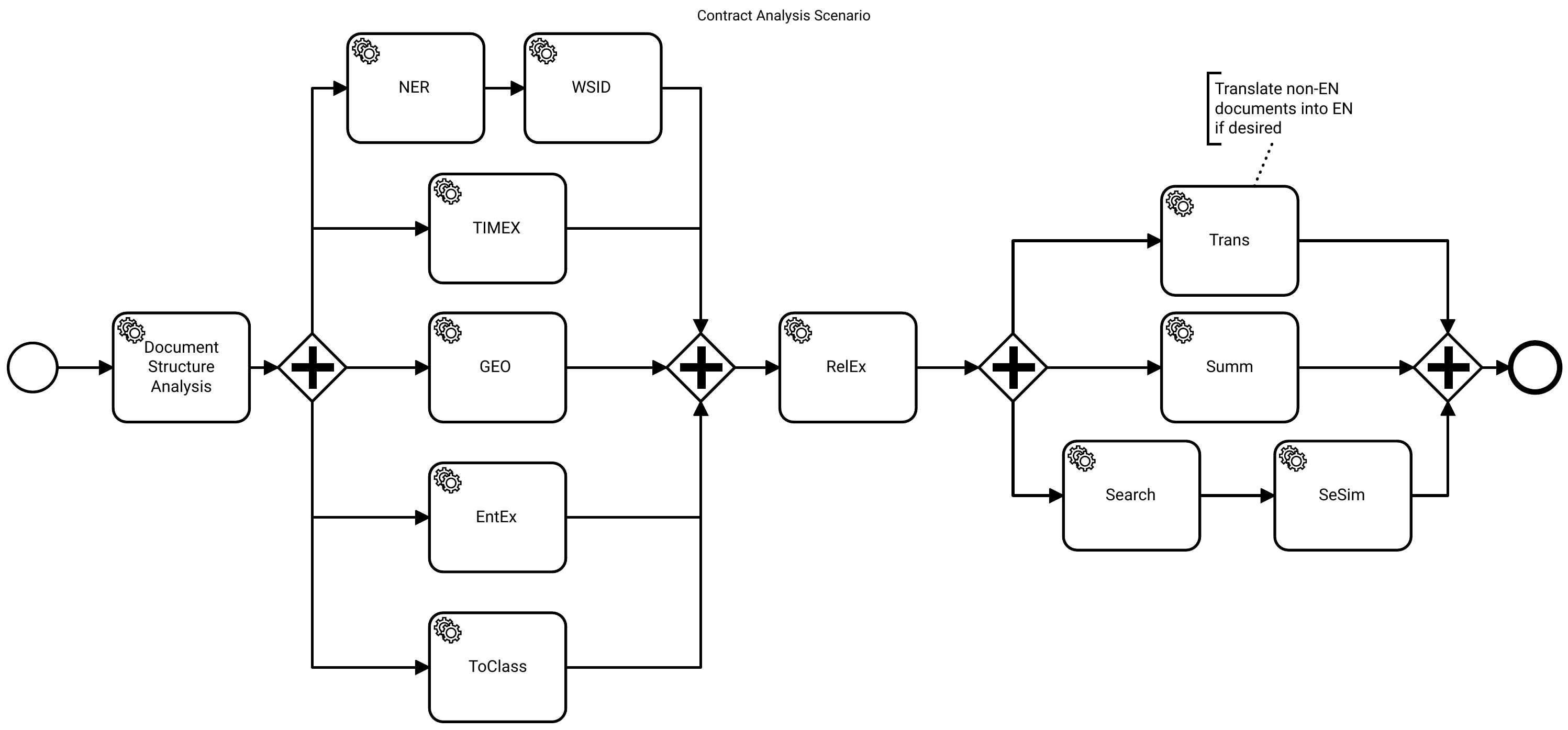}
\caption{Contract analysis workflow}
\label{fig:workflow-2}
\end{figure*}

We conceptualise the requirements of the different use cases as content curation workflows \cite{rehm2018f,rehm2016j,rehm2016p,rehm2017m}. Workflows are defined as the execution of specific services to perform the processing of one or more documents under the umbrella of a certain task or use case. The specification of a workflow includes its input and output as well as the functionality it is supposed to perform: annotate or enrich a document, add a document to the knowledge base, search for information, etc.  A workflow makes use of one or more service to implement a required functionality. For the definition of the workflows we performed a systematic analysis of the services, developed in parallel, and matched them with the required functionalities. First, we determine the principal elements involved, i.\,e., the services, input and output. Second, we define the order in which the services have to be executed. Third, we identify the components shared between workflows. 

Currently we have defined three different workflows: (1) \emph{Legal Knowledge Graph Population} is responsible for the initial population of the LKG (Figure~\ref{fig:workflow-1}) by semantically annotating and then storing documents in the LKG; (2) \emph{Contract Analysis} for the processing, analysis and enrichment of contracts (Figure~\ref{fig:workflow-2}); (3) \emph{Geothermal Project Analysis} is responsible for the analysis of geothermal project proposals in order to check their compliance with the applicable regulations. 

\section{Related Work}
\label{sec:relatedwork}

There are several systems, platforms and approaches that are related to the technology platform, which is under development in our project. In the wider area of legal document processing, technologies from several fields are relevant, among others, knowledge technologies, citation analysis, argument mining, reasoning and information retrieval. Literature overviews can be found in \cite{rehm2018g} and \cite{agnoloni2018}. 

\emph{Commercial Systems and Services --} The LexisNexis system is the market leader in the legal domain; it offers services, such as legal research, practical guidance, company research and media-monitoring as well as compliance and due diligence. WestLaw is an online service that allows legal professionals to find and consult relevant legal information.\footnote{\url{http://legalsolutions.thomsonreuters.com/law-products/westlaw-legal-research/}} One of its goals is to enable professionals to put together a strong argument. There are also smaller companies that offer legal research solutions and analytic environments, such as RavelLax,\footnote{\url{http://ravellaw.com}} or Lereto\footnote{\url{https://www.lereto.at}}. A commercial search engine for legal documents, iSearch, is a service offered by LegitQuest.\footnote{\url{https://www.legitquest.com}} The Casetext CARA Research Suite allows uploading a brief and then retrieving, based on its contents, useful case law.\footnote{\url{https://casetext.com}} There is also a growing number of startup companies active in the legal domain.
All these systems are commercial, therefore you have to pay for their use. Our platform, on the other hand, does not have a commercial base and you do not have to pay for it, only some of the services that are available under specific licenses, for which you would have to pay.

\emph{Research Prototypes --} Most of the documented research prototypes were developed in the 1990s under the umbrella of Computer Assisted Legal Research (CALR) \cite{Span1994}. In the following we briefly review several of these systems, which usually focus on one very specific feature or functionality. One example is the open source software for the analysis and visualisation of networks of Dutch case law \cite{Kuppevelt2017}. This technology determines relevant precedents (analysing the citation network of case law), compares them with those identified in the literature, and determines clusters of related cases. A similar prototype is described by \cite{Agnoloni2017}. \cite{Gifford2017} propose a search engine for legal documents where arguments are extracted from appellate cases and are accessible through selecting nodes in a litigation issue ontology or relational keyword search. Lucem \cite{Bhullar2016} mirrors the way lawyers approach legal research, developing visualisations that provide lawyers with an additional tool to approach their research results. The Eunomos  prototype semi-automates the construction and analysis of knowledge \cite{Boella2012}. 
The main difference between all these tools and our platform is the type of documents they work with. Most of these systems are limited to a single type of document, while we work with a wide variety, from contracts or laws (labour law) to industrial standards. In addition, each of these tools has a specific functionality, while the Lynx platform combines them all in a single ecosystem.

\section{Summary and Future Work}
\label{sec:conclusions}

We present the technology platform currently under development in the project Lynx, focusing upon curation workflows and processing services. These serve two main purposes: 1) to extract semantic information from large and heterogeneous sets of documents to ingest the extracted information into the Legal Knowledge Graph; 2) to extract semantic information from documents that users of the platform work with. In addition to the semantic extraction, we provide services for the processing and curation of whole documents with the goal of mapping extracted terms and concepts to the LKG, and services that aim at accessing the LKG (question answering). The final prototypes and the whole platform will be available during the last months of the project, starting from summer 2020.

Future work includes the completion of  service development, adapting the services to all languages required in the project's use cases, implementing the pilot applications and developing the web interface of the platform. In addition, we will finalise, deploy and evaluate the workflow manager and workflows defined in the project. This will not only improve the performance of the system but also simplify the way users can access the system. Last but not least, as an additional exploitation option we will explore the integration and deployment of the Lynx services through the European Language Grid \cite{elg2020}.

While some of the technologies and data sets developed in Lynx are proprietary, the following will be made openly available at the end of the project at the very latest: Legal NER for German (models and data sets) \cite{leitner2020}, Temporal Expression Analyzer for Spanish and English\footnote{\url{https://github.com/mnavasloro/Annotador}} \cite{NavasLoro2017MiningRA,navasloro2019}, Word Sense Induction and Disambiguation for English\footnote{\url{https://github.com/semantic-web-company/ptlm_wsid}} \cite{Revenko2017DiscriminationOW}, the WME and the DCM among others. 

\section*{Acknowledgments}

This work has been partially funded by the project LYNX, which has received funding from the EU's Horizon 2020 research and innovation programme under grant agreement no.~780602, see \url{http://www.lynx-project.eu}.

\section{References}
\label{main:ref}

\bibliographystyle{./lrec}
\bibliography{./lrec2020}

\end{document}